\documentclass{article}

\usepackage[preprint]{neurips_2022}

\usepackage[utf8]{inputenc} 
\usepackage[T1]{fontenc}    
\usepackage{hyperref}       
\usepackage{url}            
\usepackage{booktabs}       
\usepackage{amsfonts}       
\usepackage{nicefrac}       
\usepackage{microtype}      
\usepackage{xcolor}         

\usepackage{graphicx}
\usepackage{amsmath}
\usepackage{amsthm}
\usepackage{algorithm}
\usepackage{algorithmic}
\usepackage{multirow}
\usepackage{comment}
\usepackage{caption}
\usepackage{subcaption}
\usepackage{wrapfig}
\usepackage{cite}

\title{Pseudo Labels Regularization for Imbalanced Partial-Label Learning}

\author{
	Mingyu Xu$^{1,2}$ \qquad Zheng Lian$^{2}$\\
	$^1$School of Artificial Intelligence, University of Chinese Academy of Sciences\\
	$^2$NLPR, Institute of Automation, Chinese Academy of Sciences \\
	\texttt{\{xumingyu2021,lianzheng2016\}@ia.ac.cn} \\
}

\begin{document}

	\maketitle
	
	\begin{abstract}
		Partial-label learning (PLL) is an important branch of weakly supervised learning where the single ground truth resides in a set of candidate labels, while the research rarely considers the label imbalance. A recent study for imbalanced partial-Label learning proposed that \emph{the combinatorial challenge of partial-label learning and long-tail learning lies in matching between a decent marginal prior distribution with drawing the pseudo labels}. However, we believe that even if the pseudo label matches the prior distribution, the tail classes will still be difficult to learn because the total weight is too small. Therefore, we propose a pseudo-label regularization technique specially designed for PLL. By punishing the pseudo labels of head classes, our method implements state-of-art under the standardized benchmarks compared to the previous PLL methods. 
	\end{abstract}

	\section{introduction}
	In the real world, a large number of data are crowdsourced to non-experts for annotation. Due to the existence of various noises, the annotators may hesitate in some labels \citep{beigman2009learning,warby2014sleep,wan2020modeling}. So we can allow the annotators to give a candidate label set. To deal with the problem, partial-label learning (PLL) has attracted significant attention from the community \citep{jin2002learning,hullermeier2006learning,cour2011learning}. PLL is an important branch of weakly supervised learning, which assumes the single ground-truth label must be in the candidate set. A plethora of methods have been developed to tackle this problem, including average-based method\citep{hullermeier2006learning,cour2009learning}, identification-based\citep{yu2016maximum,lv2020progressive,feng2020learning,wen2021leveraged,wang2022pico,wang2022picoplus}, etc.
	
	However, existing PLL methods usually considered on the data set of balanced category, which may not hold in practice. In many real world scenarios, training data exhibits a long-tailed label distribution. That is, many labels occur infrequently in the training data\citep{zhang2021deep}. Unbalanced data will cause the predicted value to be biased away from the tail class\citep{menon2020long}. Recently, researchers propose a novel framework for long-tailed partial-label learning, called Solar \citep{wang2022solar}, which thinks the combinatorial challenge of partial-labeling and long-tail learning lies in matching between a decent marginal prior distribution with drawing the pseudo labels. However, we found that there is still a good result in the case of mismatching, which makes us wonder whether matching is necessary. On the other hand, the calculation of pseudo label in Solar requires Sinkhorn-Knopp algorithm \citep{cuturi2013sinkhorn}.  It is an iterative algorithm that consumes more time and space. Its convergence does not exist under the setting of PLL. For this reason, Solar also needs to use the relaxed solution occasionally. This inspired us to propose a simpler yet more effective algorithm. 
	
	Therefore, we propose a pseudo-label regularization technique for imbalanced
	partial-label learning. We do not require the matching between the pseudo labels and distribution matching and the closed-form solution of the pseudo labels can be obtained directly by using the Lagrange multiplier method. We comprehensively evaluate our method on various benchmark datasets, where our method establishes state-of-the-art performance. Compared to Solar, our method improves the accuracy on tail classes by 21.82\% on the long-tailed version of the CIFAR10 dataset. 
	
	\section{Background}
	In this section, we will first formulate the problem. Then we describe our motivation, which is derived from the classical PLL method PRODEN \citep{lv2020progressive} and the recent Long-Tail PLL method Solar\citep{wang2022solar}.
	\subsection{Problem Setup}
	Let $\mathcal{X}$ be the input space and $\mathcal{Y} = \{1,2,\cdots,c\}$ be the label space with $c$ distinct categories. We consider a partially labeled dataset $\mathcal{D}=\{(x_i, S(x_i))\}_{i=1}^{N}$ where $N$ is the number of samples and $S(x_i) \in \{0, 1\}^c$ is the candidate set for the sample $x_i \in \mathcal{X}$. We denote the $j^{th}$ element of $S(x_i)$ as $S_j(x_i)$. Here, $S_j(x_i)$ is equal to 1 if the label $j$ is a candidate label for $x_i$, and otherwise 0.
	
	Our goal is to train a classifier $f: \mathcal{X} \mapsto [0, 1]^c$, parameterized by $\theta$, that can perform predictions on
	unseen testing data. Here, $f$ is the softmax output of a neural network, and $f_j (\cdot)$ denotes the $j^{th}$ entry. To perform label disambiguation, we maintain a pseudo-label $w(x_i)$ for sample $x_i$, where $w_j(x_i)$ donate the $j^{th}$ entry. We train the classifier with the cross-entropy loss  $\sum_{j=1}^x -w_j(x_i) log f_j(x_i)$. For the convenience of writing, also record $w_j(x_i)$ as $w_{ij}$, $f_j(x_i)$ as $f_{ij}$, $S_j(x_i)$ as $S_{ij}$. 
	\subsection{Motivation}
	PRODEN is the classic way to use pseudo labels for PLL, which does not take into account the imbalance. It can be modeled as:
	\begin{align} \label{proden}
		\min_{w,\theta} &\sum_{i=1}^N \sum_{j=1}^c (-w_{ij} log f_{ij} + \frac{1}{\lambda}w_{ij} log w_{ij}) \nonumber \\
		s.t. &w_{ij} = 0\ for\ S_{ij}=0, i\in \{1,...,N\},  j\in \{1,...,c\} \nonumber\\
		&\sum_{j=1}^c w_{ij}=1, i\in \{1,...,N\}
	\end{align}
	where $\lambda =1$. To solve above problem, we can use the method of alternately optimizing $w$ and $\theta$, where we use the optimal solution $ S_{ij}f_{ij}/\sum_{j=1}^c S_{ij}f_{ij}$ to update $w_{ij}$  and use the SGD optimizer to update $\theta$.  
	
	Recently, researchers have proposed a method that can conduct long tail PLL learning, named Solar. It can be modeled as: 
	\begin{align} \label{solar}
		\min_{w,\theta} &\sum_{i=1}^N \sum_{j=1}^c (-w_{ij} log f_{ij} + \frac{1}{\lambda} w_{ij} log w_{ij}) \nonumber \\
		s.t. &w_{ij} = 0\ for\ S_{ij}=0, i\in \{1,...,N\},  j\in \{1,...,c\} \nonumber\\
		&\sum_{j=1}^c w_{ij}=1, i\in \{1,...,N\} \nonumber\\
		&\sum_{i=1}^N w_{ij}=Nr_j,j\in \{1,...,c\}
	\end{align}
	where $r_j$ donates the prior distribution of $j^{th}$ class. To solve above problem, Solar also uses the method of alternately optimizing $w$ and $\theta$. Unfortunately, this model can't write a closed-form solution like PRODEN. Therefore, Solar adopts the Sinkhorn-Knopp algorithm to obtain $w$ iteratively. However, it is difficult to guarantee the convergence of Sinkhorn-Knopp algorithm when we solve the problem \eqref{solar}. So the relaxed solution is used in Solar. In order to avoid the discussion of convergence and save time and space, \emph{can we propose a method that can directly write the closed-form solution like PRODEN for imbalanced PLL}? 
	
	Besides, we note that in Solar, the imprecise $r$ and real $r$ can lead to similar performance. This can be explained as the robustness of Solar's estimation of $r$. However, from the other side, it will also shake the argument in Solar, which is \emph{the combinatorial challenge of PLL and long-tail learning lies in matching between a decent marginal prior distribution with drawing the pseudo labels}.  And the argument leads to the constraint different from Proden in Equation \ref{proden}, $\sum_{i=1}^N w_{ij}=Nr_j$. So different from Solar, we think \emph{even if there is no match, we can also achieve good results in imbalance PLL as long as we punish the pseudo labels of head classes}. 
	
	\section{Method}
	\begin{algorithm}[t]
		\caption{Pseudo-code of Pseudo Labels Regularization for Imbalanced
			Partial-Label Learning}
		\label{alg:AOA}
		\renewcommand{\algorithmicrequire}{\textbf{Input:}}
		\renewcommand{\algorithmicensure}{\textbf{Output:}}
		\begin{algorithmic}[1]
			\REQUIRE Training dataset $\mathcal{D}$, classifier $f$, uniform marginal $r$,hyperparameters  $\lambda$, $M$.
			
			\FOR{epoch = 1,2,...}
			\FOR{step = 1,2,...}
			\STATE Get classifier prediction $f_{ij}$ on a mini-batch of data $B$.
			\STATE Get the pseudo labels by Equaion \eqref{closed-form}.
			\STATE Select sample with small loss in each category.
			\STATE Calculate classification loss, consistency loss and mixup loss.
			\STATE Update SGD optimizer to update $f$.
			\ENDFOR
			\STATE Update the marginal $r$ with the information of the training set.
			\ENDFOR
		\end{algorithmic}
	\end{algorithm}
	
	In this section, we  describe our novel framework for partial-label
	learning. This framework will meet the two core points we mentioned above, one is the method that can directly write the closed-form solution, the other is that the method can punish the pseudo labels of head classes. Then we introduce the  method of estimating prior $r$. Lastly we introduce some technologies that can improve performance, such as sample selection, consistency regularization and mixup.
	
	\subsection{Pseudo labels regularization}
	We formalize our method as:
	\begin{align} \label{PLR}
		\min_{w,\theta} &\sum_{i=1}^N \sum_{j=1}^c (-w_{ij} log f_{ij} + \frac{1}{\lambda} w_{ij} log w_{ij} + \frac{M}{\lambda}w_{ij} log r_j ) \nonumber \\
		s.t. &w_{ij} = 0\ for\ S_{ij}=0, i\in \{1,...,N\},  j\in \{1,...,c\} \nonumber\\
		&\sum_{j=1}^c w_{ij}=1, i\in \{1,...,N\} \nonumber\\
	\end{align}
	while $\lambda,M>0$. The objective function consists of three items. The first item is the commonly used cross-entropy classification loss. The second item is about the entropy regularization item of the pseudo labels $w$ to avoid overconfidence. The third item is about the regularization item of the category, which will keep the pseudo labels $w$ away from the prior distribution $r$.

	Similar to PRODEN, we adopt the method of alternately optimizing $w$ and $\theta$. For constraint $w_{ij}=0 \ for\ S_{ij}=0$, we can delete 
	the items related to $w_{ij}$ in the optimization goal if $S_{ij}=0$. Then we can calculate the Hession matrix of the optimization objective as $\frac{1}{\lambda} diag(\{w_{ij}\}_{i\in \{1,...,N\},  j\in \{1,...,c\},S_{ij}=1})$, which is a positive definite matrix. This means that the objective function is a convex optimization problem with respect to $w$ and \emph{there is a unique optimal solution about $w$ in the case of fixed $\theta$}. 
	
	Using Lagrange multiplier method,
	\begin{align}
		L(w) = \sum_{i=1}^N (\sum_{j=1}^c (-w_{ij} log f_{ij} + \frac{1}{\lambda} w_{ij} log w_{ij} + \frac{M}{\lambda}w_{ij} log r_j ) -v_i (\sum_{j=1}^c w_{ij} -1))
	\end{align}
	Let $\nabla_{w} L=0$, we can get:
	\begin{align}
		-log f_{ij} +\frac{1}{\lambda}(logw_{ij}+1) + \frac{M}{\lambda}logr_j - v_i = 0 , \ if\ S_{ij} =1
	\end{align}
	which means:
	\begin{align}
		w_{ij} = \frac{f_{ij}^{\lambda}r_{j}^{-M}}{e^{-\lambda+v_i}}, \ if\ S_{ij}=1
	\end{align}
	We can rewrite it as $w_{ij} = \frac{S_{ij}f_{ij}^{\lambda}r_{j}^{-M}}{e^{-\lambda+v_i}}$ whether $S_{ij}=0$ or $S_{ij}=1$. Combined with the constraint $\sum_{j=1}^c w_{ij} = 1$, we can get 
	\begin{align}
		\sum_{j=1}^c \frac{S_{ij}f_{ij}^{\lambda}r_{j}^{-M}}{e^{-\lambda+v_i}} = \sum_{j=1}^c w_{ij}=1 \nonumber \\
		e^{-\lambda+v_i} = \sum_{j=1}^cS_{ij}f_{ij}^{\lambda}r_j^{-M}
	\end{align}
	
	and the optimal $w$ satisfy:
	\begin{align}\label{closed-form}
		w_{ij} = \frac{S_{ij}f_{ij}^{\lambda}r_j^{-M}}{\sum_{j=1}^cS_{ij}f_{ij}^{\lambda}r_j^{-M}}
	\end{align}
	Besides, we use the SGD optimization to update $\theta$.
	
	Because we can directly write the optimal closed-form solution of $w$ in each iteration process, the cost of space and time is very small. And \emph{our method is consistent with Proden under the condition of balanced PLL}.  This is because when $r_1=r_2=...=r_c$, Equation \eqref{closed-form} degenerate to PRODEN. And when $M>0$, the pseudo label of head class is punish hardly by big $r_j^{-M}$. Because head class has many samples, being slightly punished will not affect its performance too much. However, tail class can benefit from equation \eqref{closed-form} whose performance will be greatly improved, so as to improve the overall performance. $M$ is a balance factor to balance head classes and tail classes training in training. Another case where our method degenerates to PRODEN is $M=0$. 
	\subsection{Other technologies}
	Pseudo labels regularization is the core component of our method. However, in order to better conduct the long-tail PLL learning, we also adopted the following technologies.
	
	\textbf{Estimate the prior distribution} There is a difference between long-tail learning and long-tail PLL in that the number of each category is unknown, which requires us to estimate $r$ based on the information of training data. We initialize $r$ to be uniformly distributed $[1/c,...,1/c]$, and update it by using a Moving-Average strategy:
	\begin{align} \label{esr}
		r_j \gets \mu r_j + (1 - \mu) \frac{1}{n} \sum_{i=1}^n 1_{j = argmax_{1\leq j \leq c} P_{ij}} 
	\end{align}
	where $\mu \in [0, 1]$ is a preset scalar. However, the way to estimate $r$ is not unique. We will also try a variety of different estimation methods in the experimental part.

	\textbf{Consistency regulation} Using strong and weak data augmentation is an effective method for PLL learning, which has been verified in PiCO \citep{wang2022pico},  CRDPLL\citep{wu2022revisiting}, and IRNet \citep{lian2022irnet}. Therefore, we also use consistency regulation in long-tail PLL. Specifically,  we calculate the cross entropy loss by using the pseudo label of weakly augmented samples and the prediction of strongly augmented samples.
	
	\textbf{Mixup}. Recently, mixup \citep{zhang2018mixup} technology has been used in PLL to enhance the robust of PLL in PiCO+ \citep{wang2022picoplus} and DALI \citep{xu2023dali}. It also shows excellent performance in the long-tail PLL in Solar\citep{wang2022solar}. In order to improve performance, we also adopted the mixup technology. That is to construct a new sample whose input is a linear combination of two samples, and its pseudo label is also a linear combination of the two samples.
	
	\textbf{Sample selection} After we get the pseudo label  from equation \eqref{PLR}, only some pseudo are actually trustworthy.  Therefore, we will select small loss samples from each type of samples to perform consistency loss and mixup, which will improve the representation ability of our method. We first calculate the samples belonging to $k^{th}$ class from the batch $B$: $B_k  = \{(x_i,w_i) \in B |k=argmax_{1 \leq j \leq c }w_{ij} \}$. Then select the small cross entropy loss  sample with numbers of $min(|B_k|,\rho r_k |B|)$  in each $B_k$, where $\rho \in [0,1]$ is a threshold hyper-parameter. Sample select can be seen as a way of curriculum learning \citep{wang2021survey}.
	
	\section{Experiments}
	In this section, we experimentally analyze the proposed method under various scenarios for the imbalanced PLL problem.
	
	\subsection{Setup}
	Datasets. First, we evaluate our method on two long-tailed datasets CIFAR10-LT and CIFAR100-LT \citep{cao2019learning,wei2021crest}. The training images are randomly removed class-wise to follow a pre-defined
	imbalance ratio $\gamma =n_1/n_L$, where $n_j$ is the image number of the $j^{th}$ class. For convenience, class
	indices are sorted based on the class-wise sample size, in descending order with $n_1 \geq ... \geq n_L$. We
	have $n_1 = 5 000$ for CIFAR10-LT and $n_1 = 500$ for CIFAR100-LT. We use different imbalance ratios to evaluate the performance of our method, with $\gamma \in \{50, 100, 200\} $ for CIFAR10-LT and $\gamma \in
	\{10, 20, 50\}$ for CIFAR100-LT. We then generate partially labeled datasets by manually flipping
	negative labels $\hat{y} \neq y$ to false-positive labels with probability $\psi = P(\hat{y}\in \mathcal{Y} |\hat{y} \neq y)$, which follows the settings in previous works \citep{wang2022solar}. The final candidate label set is composed of the ground-truth label and the flipped false-positive labels. We choose $\psi \in \{0.3 ,0.5 \}$ { for CIFAR10-LT and  $\psi \in \{0.05 ,0.1 \}$ for CIFAR100-LT. For all experiments, we report the mean and standard deviation based on 3 independent runs (with different random seeds).
		
		\textbf{Baselines}. We compare our method with seven state-of-the-art partial-label learning methods: 1) PiCO \citep{wang2022pico} leverages contrastive learning to disambiguate the candidate labels by updating the pseudo-labels
		with contrastive prototype labels. 2) PRODEN  \citep{lv2020progressive} is also a pseudo-labeling method that iteratively updates the latent label distribution by re-normalized classifier prediction. 3) VALEN \citep{xu2021instance}  recovers the latent label distributions by a Bayesian parametrization model. 4) LWS \citep{wen2021leveraged} also works in a pseudo-labeling style, which weights the risk function by considering the trade-off between losses on candidate labels and non-candidate ones. 5) CC \citep{feng2020provably} is a classifier-consistent method that assumes the candidate label set is uniformly sampled. 6) MSE and EXP \citep{feng2020learning} utilize mean square error and exponential loss as the risk estimators. All the hyper-parameters are searched according to the original papers. 7)  Solar is a long-tail PLL method to match pseudo labels with prior. 
		
		\textbf{Implementation details}. We use an 18-layer ResNet as the feature backbone. The model is trained for 1000 epochs using a standard SGD optimizer with a momentum of 0.9. The initial learning rate is set as 0.01, and decays by the cosine learning rate schedule. The batch size is 256. These configurations are applied for our method and all baselines for fair comparisons. We devise
		a pre-estimation training stage, where we run a model for 100/20 epochs (on CIFAR10/100-LT) respectively to obtain a coarse-grained class prior, which is the same with the previous works \citep{wang2022solar}. After that, we re-initialize the model weights and
		run with this class prior.  We use $\lambda =3,M=2$ for CIFAR10-LT, and $\lambda =1,M=0.5$ for CIFAR100-LT. The moving-average parameter $\mu$ for class prior estimation is set as 0.1/0.05 in the first stage and fixed as 0.01
		later. For class-wise reliable sample selection, we linearly ramp up $\rho$ from 0.2 to 0.5/0.6 in the first
		50 epochs.  For fair comparisons, we equip all the baselines
		except PiCO with consistence loss and mixup. The mix coefficient of mixup is sampled from $beta(4,4)$. 
		\begin{table}[t]
			\caption{Accuracy comparisons on CIFAR10-LT and CIFAR100-LT under various flipping probability
				$\psi$ and imbalance ratio $\gamma$. Bold indicates superior results.} \label{tabel1}
			\centering
			\begin{tabular}{c|ccc|ccc} 
				\hline
				\multirow{3}{*}{methods} & \multicolumn{6}{c}{CIFAR10-LT}\\ 
				\cline{2-7} & \multicolumn{3}{c|}{ $\psi=0.3$} & \multicolumn{3}{c}{ $\psi=0.5$}  \\ 
				\cline{2-7}&  $\gamma=50$ & $\gamma=100$ & $\gamma=200$ &  $\gamma=50$ &  $\gamma=100$ &  $\gamma=200$  \\ 
				\hline
				MSE & 61.13±1.08 & 52.59±0.48& 48.09±0.45  & 49.61±1.42 & 43.90±0.77& 39.52±0.70\\
				EXP & 52.93±3.44 & 43.59±0.16& 42.56±0.44& 50.62±3.00 & 43.69±2.72& 41.07±0.62\\
				LWS & 44.51±0.03 & 43.60±0.12& 42.33±0.58   & 24.62±9.67& 27.33±1.84& 28.74±1.86\\
				VALEN & 58.34±1.05  & 50.20±6.55& 46.98±1.24& 40.04±1.88  & 37.10±0.88& 36.61±0.57  \\
				CC  & 78.76±0.27 & 71.86±0.78& 63.38±0.79& 73.09±0.40 & 64.88±1.03 & 54.41±0.85\\
				PRODEN & 81.95±0.19& 71.09±0.54&63.00±0.54& 66.00±3.60&62.17±3.36 & 54,65±1.00 \\
				PiCO & 75.42±0.49   & 67.73±0.64              & 61.12±0.67 & 72.33±0.08  & 63.25±0.64& 53.92±1.64   \\
				Solar& 83.80±0.52 & 76.64±1.66& 67.47±1.05 & 81.38±2.84 & 74.16±3.03& 53.92±1.64\\
				Ours& \textbf{87.25}±0.51 & \textbf{81.74}±0.53 & \textbf{74.07}±1.45& \textbf{85.86}±1.01  & \textbf{78.38}±0.37 & \textbf{65.76}±2.86   \\
				\hline
			\end{tabular}
			\begin{tabular}{c|ccc|ccc} 
				\hline
				\multirow{3}{*}{methods} & \multicolumn{6}{c}{CIFAR100-LT}\\ 
				\cline{2-7} & \multicolumn{3}{c|}{ $\psi=0.05$} & \multicolumn{3}{c}{ $\psi=0.1$}  \\ 
				\cline{2-7}&  $\gamma=10$ & $\gamma=20$ & $\gamma=50$ &  $\gamma=10$ &  $\gamma=20$ &  $\gamma=50$  \\ 
				\hline
				MSE & 49.92±0.64 & 43.94±0.86 & 37.77±0.40& 42.99±0.47 & 37.19±0.72 & 31.49±0.35\\
				EXP & 25.86±0.94& 24.84±0.40& 23.58±0.47& 24.82±1.41& 21.27±1.24& 19.88±0.43\\
				LWS & 48.85±2.16& 35.88±1.29& 19.22±8.56& 6.10±2.05& 7.16±2.03& 5.15±0.36\\
				VALEN & 49.12±0.58& 42.05±1.52& 35.62±0.43& 33.39±0.65& 30.67±0.11& 24.93±0.87  \\
				CC  & 60.36±0.52& 54.33±0.21& 45.83±0.31& 57.91±0.41& 51.09±0.48& 41.74±0.41\\
				PRODEN & 60.31±0.50& 50.39±0.96& 42.29±0.44& 47.32±0.60& 41.82±0.55& 35.11±0.08 \\
				PiCO & 54.05±0.37& 46.93±0.65& 38.74±0.11& 46.49±0.46& 39.80±0.34& 34.97±0.09
				\\
				Solar& 64.75±0.71& 56.47±0.76& 46.18±0.85& 61.82±0.71& 53.03±0.56& 40.96±1.01\\
				Ours& \textbf{65.83±0.43} & \textbf{58.62}±0.61 & \textbf{48.73}±0.25& \textbf{63.89}±0.63  & \textbf{54.49}±0.64 & \textbf{45.74}±0.70   \\
				\hline
			\end{tabular}
		\end{table}
		\begin{table}
			\caption{Results on different groups of labels  comparisons on CIFAR10-LT ($\psi= 0.5, \gamma = 100$) and CIFAR100-
				LT ($\psi = 0.1, \gamma = 20$). The best results are marked in bold and the second-best marked in underline.} \label{tabel2}
			\centering
			\begin{tabular}{c|cccc|cccc} 
				\hline
				\multirow{2}{*}{methods}  & \multicolumn{4}{c|}{ CIFAR10-LT} & \multicolumn{4}{c}{ CIFAR100-LT}  \\ 
				\cline{2-9}&  ALL &Many &Medium& Few& ALL &Many& Medium& Few \\ 
				\hline
				MSE &43.90 & 81.11& 42.03& 9.18& 37.19 &57.46& 37.57& 16.53\\
				EXP &43.69 & 92.64& 39.75& 0.00& 21.27 &60.35& 3.99& 0.00\\
				LWS &27.33& 89.09& 1.52& 0.00& 7.16&20.80& 0.86& 0.00\\
				VALEN &37.10& 85.30& 28.78& 0.00& 30.67&58.74& 16.25& 0.07  \\
				CC  &64.88& 94.56& 65.61& 34.22& 51.09&73.03& 52.15& 28.05\\
				PRODEN &62.17& \textbf{96.83}& 72.18& 14.17&41.82& \textbf{76.86}& 43.14& 5.43 \\
				PiCO &63.25& 93.33& 66.14& 29.30& 39.80&70.75& 42.42& 6.14   \\
				Solar& \underline{76.16}&\underline{96.50}& \underline{76.01}& \underline{49.34}& \underline{53.03}&74.33& 54.09& \textbf{30.62}\\
				Ours&\textbf{78.38}& 85.11& \textbf{78.75} &\textbf{71.16} &\textbf{54.49}&\underline{74.59}&\textbf{58.43}&\underline{30.19}   \\
				\hline
			\end{tabular}
		\end{table}
		\subsection{Main result}
		
		\textbf{Our method achieves SOTA results}. As shown in Table \ref{tabel1}, our method significantly outperforms all the rivals by a notable margin under various settings of imbalance ratio $\gamma$ and label ambiguity degree $\psi$.
		Specifically, on CIFAR10-LT dataset with $\psi = 0.5$ and  the imbalance ratio $\gamma=200$ we improve upon the best baseline by 11.84\%. Specifically, on CIFAR100-LT dataset with $\psi$ = 0.5 and the imbalance ratio $\gamma=50$, we improve upon the best baseline by 4.00\%. Our method is superior to previous methods in all cases. Especially with the increase of the imbalance rate, our method can still show excellent performance.
		
		\textbf{Results on different groups of labels}. We show that our method achieves overall strong performance on both medium and tail classes. To see this, we report accuracy on three groups of classes with different sample sizes. Recall from Section 4.1 that the class indices are sorted based on the sample size, in
		descending order. We divide the classes into three groups: many ({1, 2, 3}), medium ({4, 5, 6, 7}), and few ({8, 9, 10}) shots for CIFAR10-LT and many ({1, . . . , 33}), medium ({34, . . . , 67}), and few ({68, . . . , 100}) shots for CIFAR100-LT. Table 2 shows the accuracy of different groups on both CIFAR10-LT and CIFAR100-LT.  The comparisons highlight that our method poses a good disambiguation ability in the long-tailed PLL setup. We can find that our method sacrifices the performance of the head classes, but improves the performance of the middle and tail classes.
		\subsection{Ablation Studies} Next, we will do ablation experiments to illustrate the impact of various technologies in Section 3.2, and can further explain the performance of our proposed methods.
		
		\begin{table}[t]
			\caption{Ablation results on CIFAR10-LT ($\psi= 0.5, \gamma = 100$) and CIFAR100-
				LT ($\psi = 0.1, \gamma = 20$).} \label{tabel3}
			\centering
			\begin{tabular}{c|cccc|cccc} 
				\hline
				\multirow{2}{*}{methods}  & \multicolumn{4}{c|}{ CIFAR10-LT} & \multicolumn{4}{c}{ CIFAR100-LT}  \\ 
				\cline{2-9}&  ALL &Many &Medium& Few&ALL& Many& Medium& Few \\ 
				\hline
				Ours & 78.38& 85.11& 78.75&71.16& 54.49&74.59&58.43&30.19\\
				Solar & 74.16& 96.50 &76.01& 49.34& 53.03& 74.33 &54.09 &30.62\\\hline
				Ours w/o S & 64.28& 87.59& 69.42& 34.11& 51.06& 76.71 &55.89 &20.42\\
				Solar w/o S & 29.61 & 86.10 & 6.95 & 3.34 & 35.80 & 66.85 & 32.06 & 8.61 \\ \hline
				Ours w/o MU & 65.27& 83.53& 62.70& 47.33& 51.97& 71.51 &55.36 &28.95\\
				Solar  w/o MU & 69.40& 92.77& 72.75& 41.58& 47.41& 71.45& 48.06& 22.70  \\ \hline
				ours  w/o CR & 73.45 &81.07& 75.38& 62.72& 51.61& 71.33& 52.96& 30.51  \\
				Solar  w/o CR & 57.97& 92.78& 61.05& 19.05& 47.74& 70.18& 51.85& 21.06  \\ \hline
				ours  w/o MU+CR & 45.97 &68.60& 41.46& 29.34& 39.83& 56.36& 41.55& 21.52  \\
				Solar  w/o MU+CR & 44.83& 82.33& 40.39& 13.25& 30.88& 50.52& 30.53& 11.61  \\
				PRODEN  w/o MU+CR & 46.61& 85.43& 44.65& 10.40& 31.78& 55.09& 32.38& 7.85  \\
				CC  w/o MU+CR & 36.98& 79.38& 30.63& 3.04& 25.96& 47.08& 23.82& 7.04  \\
				\hline
			\end{tabular}
		\end{table}

		\textbf{Sample select or not.} Firstly, we experiment with sample select or not. As a comparison, we also conducted the desired ablation experiment on Solar. w/o S means regards all examples as clean samples and does not perform selection. The results are shown in Table \ref{tabel3}. We can find that sample select has improved our method. When we do not use sample select, our method improves by 34.67\% in CIFAR10-LT and 15.26 \% in CIFAR100-LT compared with Solar. Our method does not rely too much on sample selection as solar. This also shows the  power of the pseudo label regularization technique we proposed for imbalance PLL. 
		
		\textbf{Consistency regularization and Mixup or Not} we ablate the contributions of two components in representation enhancement: mixup augmentation training and consistency regularization. w/o MU which removing Mixup augmentation
		training. w/o CR means  removing consistency regularization. w/o MU+CR means
		removing both Mixup and consistency. The results are shown in Table \ref{tabel3}. 
		We can find that all methods can benefit from consistency regularization and mixup. But our methods and Solar benefit the most.  In the eight groups experiments shown in Table \ref{tabel3}, the performance of our method is the first in most cases.
		\subsection{Further Analysis} 
		
		\begin{table}[t]
			\caption{Method of estimating $r$ on CIFAR10-LT ($\psi= 0.5, \gamma = 100$) and CIFAR100-
				LT ($\psi = 0.1, \gamma = 20$).} \label{tabel5}
			\centering
			\begin{tabular}{c|cccc|cccc} 
				\hline
				\multirow{2}{*}{methods}  & \multicolumn{4}{c|}{ CIFAR10-LT} & \multicolumn{4}{c}{ CIFAR100-LT}  \\ 
				\cline{2-9}&  ALL &Many &Medium& Few&ALL& Many& Medium& Few \\ 
				\hline
				formula \eqref{esr} & 78.38& 85.11& 78.75&71.16& 54.49&74.59&58.43&30.19\\
				formula \eqref{esr2} & 79.51& 85.23 &78.90& 74.60& 49.70& 71.21 &52.65&25.15\\
				formula \eqref{esr3}  & 72.36& 79.93& 75.12& 61.10& 52.60& 68.85 &57.65&31.15\\
				\hline
			\end{tabular}
		\end{table}
		\textbf{Method of estimating $r$}. In section 3.2, we use formula \eqref{esr} to estimate $r$. But this is only heuristic. We can try other ways, such as: 
		\begin{align} \label{esr2}
			r_j \gets \mu r_j + (1 - \mu) \frac{1}{n} \sum_{i=1}^n P_{ij}
		\end{align}
		\begin{align} \label{esr3}
			r_j \gets \mu r_j + (1 - \mu) \frac{1}{n} \sum_{i=1}^n 1_{j = argmax_{1\leq j \leq c} w_{ij}} 
		\end{align}

		The results are shown in Table \ref{tabel5}. We can find in some case, formula \eqref{esr2} performance better than the default case.  We can also find that using prediction information is better than using pseudo label information to estimate $r$. 
		
		\textbf{Influence of regularization coefficient $\lambda$ and $M$}. In this section, we will try different regularization coefficients. We conduct experiments on CIFAR10-LT with $\psi=0.5$ and $\gamma=100$. The result are show in Table 6. We find that selecting appropriate regularization coefficient is the key to improve the performance of Long-tail PLL. And we find that with the increase of $\lambda$, the optimal $M/ \lambda$ also increases, where $M/\lambda$ is the regularization coefficient in formula \eqref{PLR}. When $\lambda=1$,$M=0$, our method can be PRODEN. Almost all results shown in Table \ref{tabel6} are better than 62.17, which is the accuracy of PRODEN in Table \ref{tabel1}
		
		\textbf{Results on fine-grained partial-label learning}. In practice, semantically similar classes can lead to
		significant label ambiguity, as exemplified in Table \ref{tabel7}. To test the limit of our method, we follow Solar \citep{wang2022solar} and evaluate on two fine-grained datasets: 1) CUB200-LT \citep{welinder2010caltech} dataset with 200 bird species; 2)
		CIFAR100-LT with hierarchical labels (CIFAR100-H-LT), where the candidate labels are generated
		within the same superclass
		. We set $\psi = 0.05, \gamma = 5$ for CUB200-LT and $\psi = 0.5$, $\gamma = 20 $ for CIFAR100-H-LT. These results clearly validate the effectiveness of our method,  when the dataset presents severe label ambiguity.
		
		\begin{table}[t]
			\centering
			\caption{Influence of regularization coefficient $\lambda$ and $M$ on CIFAR10-LT ($\psi= 0.5, \gamma = 100$).} \label{tabel6}
			\begin{minipage}{0.495\linewidth}
				\caption*{$\lambda=4$}
				\begin{tabular}{c|cccc}\hline
					M & All   & Many  & Medium & Few    \\ \hline
					1     & 72.04 & 89.67 & 59.92  & 70.59  \\
					2     & 76.15 & 87.32 & 77.77  & 62.82  \\
					3     & 76.61 & 84.18 & 79.39  & 65.32  \\
					4     & 70.62 & 82.11 & 69.68  & 60.4  \\\hline
				\end{tabular}
				
			\end{minipage}
			\begin{minipage}{0.495\linewidth}
				\caption*{$\lambda=3$}
				\begin{tabular}{c|cccc}\hline
					M & All   & Many  & Medium & Few    \\ 
					\hline
					1     & 73.42 & 88.65 & 67.96  & 65.48  \\
					1.5   & 77.49 & 87.07 & 78.33  & 66.89  \\
					2     & 78.38 & 85.11 & 78.75  & 71.16  \\
					3     & 68.55 & 79.31 & 71.23  & 54.29  \\
					\hline
				\end{tabular}
				
			\end{minipage}
			
			\begin{minipage}{0.495\linewidth}
				\caption*{$\lambda=2$}
				\begin{tabular}{c|cccc}\hline
					M& All   & Many  & Medium & Few    \\ 
					\hline
					0.5   & 71.33 & 90    & 67.25  & 58.09  \\
					1     & 77.49 & 86.66 & 79.17  & 66.07  \\
					1.5   & 73.24 & 82.81 & 78.87  & 56.18  \\
					2     & 68.1  & 75.38 & 74     & 52.95  \\
					\hline
				\end{tabular}
				
			\end{minipage}
			\begin{minipage}{0.495\linewidth}
				\caption*{$\lambda=1$}
				\begin{tabular}{c|cccc}\hline
					M & All   & Many  & Medium & Few    \\ 
					\hline
					0.25  & 72.94 & 86.74 & 71.11  & 61.59  \\
					0.5   & 67.24 & 64.99 & 74.38  & 59.93  \\
					0.75  & 65.44 & 67.28 & 71     & 51.91  \\
					1     & 61.35 & 56.47 & 68.38  & 54.76  \\
					\hline
				\end{tabular}
			\end{minipage}
		\end{table}
		
		\begin{table}[t]
			\caption{Performance comparisons of our method
				and Solar on the fine-grained CUB200-LT dataset and on
				the CIFAR100-LT dataset with hierarchical labels (CIFAR100-H-LT)} \label{tabel7}
			\centering
			\begin{tabular}{c|cccc|cccc} 
				\hline
				\multirow{2}{*}{methods}  & \multicolumn{4}{c|}{ CUB200} & \multicolumn{4}{c}{ CIFAR100-H-LT}  \\ 
				\cline{2-9}&  ALL &Many &Medium& Few&ALL& Many& Medium& Few \\ 
				\hline
				Ours & 39.98& 61.19& 44.31&15.79& 58.88&76.06&60.33&40.17\\
				Solar & 38.96& 58.93 &42.44& 17.18& 58.09& 76.78 &58.26&39.13\\
				\hline
			\end{tabular}
		\end{table}

		\begin{table}[t]
			\caption{Performance comparisons of our method
				and logits adjustment on the CIFAR1-LT($\psi = 0.5$,$\gamma=100$) dataset and on
				the CIFAR100-LT dataset($\psi=0.1$,$\gamma = 20$).} \label{tabel8}
			\centering
			\begin{tabular}{c|cccc|cccc} 
				\hline
				\multirow{2}{*}{methods}  & \multicolumn{4}{c|}{CIFAR10-LT} & \multicolumn{4}{c}{ CIFAR100-LT}  \\ 
				\cline{2-9}&  ALL &Many &Medium& Few&ALL& Many& Medium& Few \\ 
				\hline
				Ours-IR & 61.33&	95.2&	81.92&	0.00 &	42.66	&77.03	&43.38&	7.55\\
				Ours & 78.38&85.11&	78.75&71.16&54.49&74.59&	58.43&30.1\\
				Ours-IR+LA($\phi=0.3$) & 61.61&	94.9&	82.68&	0.23&	43.09&	76.91&	44.21&	8.12
				\\
				Ours-IR+LA($\phi=0.5$) & 64.38&	84.67&	78.57&	25.17&	43.03	&76.21	&44.29&	8.55
				\\
				Ours-IR+LA($\phi=0.7$) & 62.48	&84.3	&74.6	&24.5	&40.92	&71.39	&42.44	&8.88
				\\
				Ours-IR+LA($\phi=1$) & 55.39&	71.37	&65.73	&25.63	&39.86	&67.18	&43.12	&9.18
				\\
				\hline
			\end{tabular}
		\end{table}

		\textbf{Comparison with logits adjustment}. Some methods suitable for long tail distribution learning are difficult to be directly applied to imbalanced PLL. We apply the Logits adjustment method to the unbalanced PLL and compare it. The difference is that $argmax_j g_j(x) - \phi log r_i$ is used for prediction, where $g(x)$ is the logits. The results are shown in Table \ref{tabel8}. Ours-IR means remove unbalanced regularization items $w_{ij} log r_j$ for  our method in formula \eqref{PLR}. Ours-IR+LA means equip Ours-IR with logits adjustment. We found that logits adjustment can improve performance by selecting appropriate $\phi$. But it is far less than our promotion.
		
		\textbf{Space complexity cost}. The space complexity cost of pseudo labels regularization  is $O(|B|c)$ in each batch, which is the same as PRODEN, where $|B|$ is the size of batch size, $c$ is the num of class. While Solar is $|Q|c$, where the default $|Q|$ is set as $|q||B|$ and $q=64$, $T$ is the iteration number of SK algorithm with $50$ as default. 
		
		\begin{wraptable}{r}{8cm}
			\centering
			\caption{The running time (in second) of one epoch. Solar,Ours, PRODEN mean the time cost of the pseudo label computing of the above methods. Model training means the  time cost of the rest part. }
			\begin{tabular}{c|cc}
				& CIFAR10-LT & CIFAR100-LT  \\
				\hline
				Solar          & 0.574      & 0.769        \\
				Ours & 0.016      & 0.023        \\
				PRODEN	& 0.009	&0.011 \\
				\hline
			\end{tabular}
		\end{wraptable}
		
		\textbf{Time complexity cost}. The time complexity cost of pseudo labels regularization  is $O(|B|c)$ in each batch, which is the same as PRODEN, where $|B|$ is the size of batch size, $c$ is the num of class. While Solar is $O(T|Q|c)$, where the default $|Q|$ is set as $|q||B|$ and $q=64$. And we test on a NVIDIA V-100 GPU to compare their time cost. The result are shown in Table 9.  Experiments have proved that our pseudo labels calculation method is efficient. It only takes less than 2\% of the whole training time.
		\section{Related works}
		
		\textbf{Partial-label Learning (PLL)} \citep{jin2002learning} allows each example to be equipped with a candidate label set with the ground-truth label being concealed. A plethora of PLL methods  have been developed \citep{cour2011learning,wang2019adaptive,gong2021generalized,lyu2022deep,li2022learning}. The most intuitive solution is the average-based method, which assumes that each candidate label has an equal probability of being the ground-truth label. Typically,  leveraged k-nearest neighbors for label disambiguation \citep{hullermeier2006learning},  maximized the average output of candidate and non-candidate labels in parametric models \citep{cour2009learning}. However, average-based methods can be severely affected by false positive labels in the candidate set. A more popular method is identification-based methods. proposed a self-training strategy that disambiguated candidate labels via model outputs.  LWS proposed a family of loss functions for label disambiguation. More recently, PiCO \citep{wang2022pico} introduced contrastive learning into PLL. However, previous work often failed to consider the label imbalance. Recently, Solar \citep{wang2022solar} was proposed for imbalanced partial label learning firstly.
		Solar uses the method of matching the pseudo labels with distribution, but we use the method of keeping the pseudo labels away from distribution.
		
		\textbf{Long-tailed Learning (LTL)} Recently, numerous works have focused on learning from the data with the imbalanced distribution. A common way is to over-sampling few-shot classes or under-sampling many-shot classes. Early work is often limited to supervising long tail learning \citep{chawla2002smote,he2009learning,buda2018systematic,byrd2019effect,menon2020long}. More recently, semi-supervised long-tailed learning has been considered \citep{kim2020distribution,wei2021crest,peng2023dynamic}. One of the first methods is called DARP \citep{kim2020distribution}, which refine raw pseudo-labels via a convex optimization. Our work is also put forward from the perspective of a convex optimization like DARP. DRw \citep{peng2023dynamic} reweight the sample with the number of each class. In our work,  the sum of pseudo labels of each sample is 1, which means the weight if each sample is the same if we don‘t adopt sample select. In the future, we will also try to give different weights to the samples, which means the constraint condition $\sum_{j=1}^c w_{ij}=1$ in Formula \eqref{PLR} can be relaxed.
		
		\section{Conclusion}
		In this work, we present a novel framework for a challenging imbalanced partial-label learning problem. Key to our method, we derive an pseudo labels regularization to away from the estimate class prior. Comprehensive experiments show that our method improves baseline algorithms by a significant margin. In the future, we will explore the optimal regularization coefficient $\lambda$ and $M$ theoretically. 
		\newpage
		\bibliography{nips}
		\bibliographystyle{nips}
	\end{document}